\documentclass{article}
\usepackage{spconf,epsfig}
\usepackage{graphicx}
\usepackage[noend]{algpseudocode}
\usepackage{multirow}
\usepackage{algorithm}
\usepackage{amssymb,amsmath,graphicx}
\usepackage{moredefs} 

\defcommand{\vec}[1]{\mathbf{#1}} 

\title{Improvements to Deep Convolutional Neural Networks for LVCSR}
\name{Tara N. Sainath$^1$, Brian Kingsbury$^1$, Abdel-rahman Mohamed$^2$, George E. Dahl$^2$, George Saon$^1$}
\nameplus{Hagen Soltau$^1$, Tomas Beran$^1$, Aleksandr Y. Aravkin$^1$, Bhuvana Ramabhadran$^1$}
\address{$^1$IBM T. J. Watson Research Center, Yorktown Heights, NY 10598\\
$^2$Department of Computer Science, University of Toronto \\
$^1$\{tsainath, bedk, gsaon, hsoltau, tberan, saravkin, bhuvana\}@us.ibm.com, \\
$^2$\{asamir, gdahl\}@cs.toronto.edu}

\begin{document}
\maketitle
\ninept
\begin{abstract}
Deep Convolutional Neural Networks (CNNs) are more powerful than Deep Neural Networks (DNN), as they are able to better reduce spectral variation in the input signal. This has also been confirmed experimentally, with CNNs showing improvements in word error rate (WER) between 4-12\% relative compared to DNNs across a variety of LVCSR tasks. In this paper, we describe different methods to further improve CNN performance. First, we conduct a deep analysis comparing limited weight sharing and full weight sharing with state-of-the-art features. Second, we apply various pooling strategies that have shown improvements in computer vision to an LVCSR speech task. Third, we introduce a method to effectively incorporate speaker adaptation, namely fMLLR, into log-mel features. Fourth, we introduce an effective strategy to use dropout during Hessian-free sequence training. We find that with these improvements, particularly with fMLLR and dropout, we are able to achieve an additional 2-3\% relative improvement in WER on a 50-hour Broadcast News task over our previous best CNN baseline. On a larger 400-hour BN task, we find an additional 4-5\% relative improvement over our previous best CNN baseline.
\end{abstract}

\section{Introduction}

Deep Neural Networks (DNNs) are now the state-of-the-art in acoustic modeling for speech recognition, showing tremendous improvements on the order of 10-30\% relative across a variety of small and large vocabulary tasks \cite{Hinton12}. Recently, deep convolutional neural networks (CNNs) \cite{lecun:cnn, abdo:cnn} have been explored as an alternative type of neural network which can reduce translational variance in the input signal. For example, in \cite{tsainath:cnnLVCSR}, deep CNNs were shown to offer a 4-12\% relative improvement over DNNs across different LVCSR tasks.  The CNN architecture proposed in \cite{tsainath:cnnLVCSR} was a somewhat vanilla architecture that had been used in computer vision for many years. The goal of this paper is to analyze and justify what is an appropriate CNN architecture for speech, and to investigate various strategies to improve CNN results further.

First, the architecture proposed in \cite{tsainath:cnnLVCSR} used multiple convolutional layers with full weight sharing (FWS), which was found to be beneficial compared to a single FWS convolutional layer. Because the locality of speech is known ahead of time, \cite{abdo:cnn} proposed the use of limited weight sharing (LWS) for CNNs in speech. While LWS has the benefit that it allows each local weight to focus on parts of the signal which are most confusable, previous work with LWS had just focused on a single LWS layer \cite{abdo:cnn}, \cite{deng:hetPool}. In this work, we do a detailed analysis and compare multiple layers of FWS and LWS. 

Second, there have been numerous improvements to CNNs in computer vision, particularly for small tasks. For example, using $l_p$ \cite{sermanet:lp} or stochastic pooling \cite{fergus:sp} provides better generalization than max pooling used in \cite{tsainath:cnnLVCSR}. Second, using overlapping pooling \cite{krizhevsky:imagenet} and pooling in time \cite{lecun:cnnVision} also improves generalization to test data. Furthermore, multi-scale CNNs \cite{sermanet:lp}, that is, combining outputs from different layers of the neural network, has also been successful in computer vision. We explore the effectiveness of these strategies for larger scale speech tasks. 

Third, we investigate using better features for CNNs. Features for CNNs must exhibit locality in time and frequency. In \cite{tsainath:cnnLVCSR} it was found that VTLN-warped log-mel features were best for CNNs. However, speaker adapted features, such as feature space maximum likelihood linear regression (fMLLR) features \cite{gales:fmllr}, typically give the best performance for DNNs. In \cite{tsainath:cnnLVCSR}, the fMLLR transformation was applied directly to a correlated VTLN-warped log-mel space. However, no improvement was observed as fMLLR transformations typically assume uncorrelated features. In this paper, we propose a methodology to effectively use fMLLR with log-mel features. This involves transforming log-mel into an uncorrelated space, applying fMLLR in this space, and then transforming the new features back to a correlated space.

Finally, we investigate the role of rectified linear units (ReLU) and dropout for Hessian-free (HF) sequence training \cite{bedk:hf} of CNNs. In \cite{gdahl:relu}, ReLU+dropout was shown to give good performance for cross-entropy (CE) trained DNNs but was not employed during HF sequence-training. However, sequence-training is critical for speech recognition performance, providing an additional relative gain of 10-15\% over a CE-trained DNN \cite{bedk:hf}.  During CE training, the dropout mask changes for each utterance. However, during HF training, we are not guaranteed to get conjugate directions if the dropout mask changes for each utterance. Therefore, in order to make dropout usable during HF, we keep the dropout mask fixed per utterance for all iterations of conjugate gradient (CG) within a single HF iteration.

Results with the proposed strategies are first explored on a 50-hr English Broadcast News (BN) task. We find that  there is no difference between LWS and FWS with multiple layers for an LVCSR task. Second, we find that various pooling strategies that gave improvements in computer vision tasks, do not help much in speech. Third, we observe that improving the CNN input features by including fMLLR gives improvements in WER. Finally, fixing the dropout mask during the CG iterations of HF lets us use dropout during HF sequence training and avoids destroying the gains from dropout accrued during CE training. Putting together improvements from fMLLR and dropout,§ we find that we are able to obtain a 2-3\% relative reduction in WER compared to the CNN system proposed in \cite{tsainath:cnnLVCSR}. In addition, on a larger 400-hr BN task, we can also achieve a 4-5\% relative improvement in WER.

The rest of this paper is organized as follows. Section 2 describes the basic CNN architecture in \cite{tsainath:cnnLVCSR} 
that serves as a starting point to the proposed modifications. In Section 3, we discuss experiments with LWS/FWS, pooling, fMLLR and ReLU+dropout for HF. Section 4 presents results with the proposed improvements on a 50 and 400-hr BN task. Finally, Section 5 concludes the paper and discusses future work.

\section{Basic CNN Architecture}

In this section, we describe the basic CNN architecture that was introduced in \cite{tsainath:cnnLVCSR}, as this will serve as the baseline system which we improve upon. In \cite{tsainath:cnnLVCSR}, it was found that having two convolutional layers and four fully connected layers, was optimal for LVCSR tasks. We found that a pooling size of 3 was appropriate for the first convolutional layer, while no pooling was used in the second layer. Furthermore, the convolutional layers had 128 and 256 feature maps respectively, while the fully connected layers had 1,024 hidden units. The optimal feature set used was VTLN-warped log-mel filterbank coefficients, including delta + double delta. Using this architecture for CNNs, we were able to achieve  between 4-12\% relative improvement over DNNs across many different LVCSR tasks.

In this paper, we explore feature, architecture and optimization strategies to improve the CNN results further. Preliminary experiments are performed on a 50-hr English Broadcast News task \cite{bedk:hf}.  The acoustic models are trained on 50 hours from the 1996 and 1997 English Broadcast News Speech Corpora. Results are reported on the EARS {\tt dev04f} set.  Unless otherwise noted, all CNNs are trained with cross-entropy, and results are reported in a hybrid setup.


\section{Analysis of Various Strategies for LVCSR}

\subsection{Optimal Feature Set}

Convolutional neural networks require features which are locally correlated in time and frequency. This implies that Linear Discriminant Analysis (LDA) features, which are very commonly used in speech, cannot be used with CNNs as they remove locality in frequency \cite{abdo:cnn}. Mel filter-bank (FB) features are one type of speech feature which exhibit this locality property \cite{ar:tSNE}. We explore if any additional transformations can be applied to these features to further improve WER. Table 3 shows the WER as a function of input feature for CNNs. The following can be observed:
\begin{itemize}
  \item Using VTLN-warping to help map features into a canonical space offers improvements.
  \item Using fMLLR to further speaker-adapt the input does not help. One reason could be that fMLLR assumes the data is well modeled by a diagonal model, which would work best with decorrelated features. However, the mel FB features are highly correlated.
  \item Using delta and double-delta (d + dd) to capture further time-dynamic information in the feature helps.
  \item Using energy does not provide improvements.
\end{itemize}

\begin{table} [h!]
  \centering
  \begin{tabular}{|c|c|}
    \hline
    Feature & WER \\  \hline
    Mel FB & 21.9 \\ \hline
    VTLN-warped mel FB& 21.3\\ \hline
    VTLN-warped mel FB + fMLLR	& 21.2\\ \hline
    VTLN-warped mel FB + d + dd& 20.7\\ \hline
    VTLN-warped mel FB + d + dd + energy &  21.0 \\ \hline
  \end{tabular}
  \caption{WER as a function of input feature}\label{table:feat}
    \vspace{-0.15in}
\end{table}

In conclusion, it appears VTLN-warped mel FB + d+dd is the optimal input feature set to use. This feature set is used for the remainder of the experiments, unless otherwise noted.

\subsection{Number of Convolutional vs. Fully Connected Layers}

Most CNN work in image recognition makes use of a few convolutional layers before having fully connected layers. The convolutional layers are meant to reduce spectral variation and model spectral correlation, while the fully connected layers aggregate the local information learned in the convolutional layers to do class discrimination. However, the CNN work done thus far in speech \cite{abdo:cnn} introduced a novel framework for modeling spectral correlations, but this framework only allowed for a single convolutional layer. We adopt a spatial modeling approach similar to the image recognition work, and explore the benefit of including multiple convolutional layers.

Table \ref{table:convLayer} shows the WER as a function of the number of convolutional and fully connected layers in the network.  Note that for each experiment, the number of parameters in the network is kept the same. The table shows that increasing the number of convolutional layers up to 2 helps, and then performance starts to deteriorate. Furthermore, we can see from the table that CNNs offer improvements over DNNs for the same input feature set.

\begin{table} [h!]
  \centering
  \begin{tabular}{|c|c|}
    \hline
    \# of Convolutional vs.  & WER \\
    Fully Connected Layers & \\ \hline
    No conv, 6 full (DNN) & 24.8 \\ \hline
    1 conv, 5 full	 & 23.5 \\ \hline
    2 conv, 4 full & 22.1\\ \hline
    3 conv, 3 full & 22.4\\ \hline
  \end{tabular}
  \caption{WER as a Function of \# of Convolutional Layers}\label{table:convLayer}
    \vspace{-0.2in}
\end{table}

\subsection{Number of Hidden Units}

CNNs explored for image recognition tasks perform weight sharing across all pixels.  Unlike images, the local behavior of speech features in low frequency is very different than features in high frequency regions. \cite{abdo:cnn} addresses this issue by limiting weight sharing to frequency components that are close to each other. In other words, low and high frequency components have different weights (i.e. filters). However, this type of approach limits adding additional convolutional layers \cite{abdo:cnn}, as filter outputs in different pooling bands are not related. We argue that we can apply weight sharing across all time and frequency components, by using a large number of hidden units compared to vision tasks in the convolutional layers to capture the differences between low and high frequency components. This type of approach allows for multiple convolutional layers, something that has thus far not been explored before in speech.

Table \ref{table:hiddenUnits} shows the WER as a function of number of hidden units for the convolutional layers. Again the total number of parameters in the network is kept constant for all experiments. We can observe that as we increase the number of hidden units up to 220, the WER steadily decreases. We do not increase the number of hidden units past 220 as this would require us to reduce the number of hidden units in the fully connected layers to be less than 1,024 in order to keep the total number of network parameters constant. We have observed that reducing the number of hidden units from 1,024 results in an increase in WER. We were able to obtain a slight improvement by using 128 hidden units for the first convolutional layer, and 256 for the second layer. This is more hidden units in the convolutional layers than are typically used for vision tasks \cite{lecun:cnn}, \cite{lecun:cnnVision}, as many hidden units are needed to capture the locality differences between different frequency regions in speech.

\begin{table} [h!]
  \centering
  \begin{tabular}{|c|c|}
    \hline
    Number of Hidden Units & WER \\  \hline
    64 & 24.1 \\ \hline
    128 & 23.0 \\ \hline
    220	& 22.1\\ \hline
    128/256 & 21.9\\ \hline
  \end{tabular}
  \caption{WER as a function of \# of hidden units}\label{table:hiddenUnits}
    \vspace{-0.2in}
\end{table}

\subsection{Limited vs. Full Weight Sharing}

In speech recognition tasks, the characteristics of the signal in low-frequency regions are very different than in high frequency regions. This allows a limited weight sharing (LWS) approach to be used for convolutional layers \cite{abdo:cnn}, where weights only span a small local region in frequency. LWS has the benefit that it allows each local weight to focus on parts of the signal which are most confusable, and perform discrimination within just that small local region. However, one of the drawbacks is that it requires setting by hand the frequency region each filter spans. Furthermore, when many LWS layers are used, this limits adding additional full-weight sharing convolutional layers, as filter outputs in different bands are not related and thus the locality constraint required for convolutional layers is not preserved. Thus, most work with LWS up to this point has looked at LWS with one layer \cite{abdo:cnn}, \cite{deng:hetPool}.

Alternatively, in \cite{tsainath:cnnLVCSR}, a full weight sharing (FWS) idea in convolutional layers was explored, similar to what was done in the image recognition community. With that approach, multiple convolutional layers were allowed and it was shown that adding additional convolutional layers was beneficial.  In addition, using a large number of hidden units in the convolutional layers better captures the differences between low and high frequency components.

Since multiple convolutional layers are critical for good performance in WER, in this paper we explore doing LWS with multiple layers. Specifically, the activations from one LWS layer have locality preserving information, and can be fed into another LWS layer.  Results comparing LWS and FWS are shown in Table \ref{table:lwsfws}.  Note these results are with stronger VTLN-warped log-mel+d+dd features, as opposed to previous LWS work which used simpler log-mel+d+dd.

For both LWS and FWS, we used 2 convolutional layers, as this was found in \cite{tsainath:cnnLVCSR} to be optimal. First, notice that as we increase the number of hidden units for FWS, there is an improvement in WER, confirming our belief that having more hidden units with FWS is important to help explain variations in frequency in the input signal. Second, we find that if we use LWS but match the number of parameters to FWS, we get very slight improvements in WER (0.1\%). It seems that both LWS and FWS offer similar performance. Because FWS is simpler to implement, as we do not have to choose filter locations for each limited weight ahead of time, we prefer to use FWS. Because FWS with 5.6M parameters (256/256 hidden units per convolution layer) gives the best tradeoff between WER and number of parameters, we use this setting for subsequent experiments.

\begin{table}[h!]
\begin{center}
\begin{tabular}{|c|c|c|c|} \hline
Method & Hidden Units in Conv Layers& Params & WER  \\ \hline
FWS & 128/256 & 5.1M &  19.3 \\ \hline
FWS & 256/256 & 5.6M & 18.9 \\ \hline
FWS & 384/384 & 7.6M & 18.7 \\ \hline
FWS & 512/512 & 10.0M & 18.5 \\ \hline \hline
LWS & 128/256 & 5.4M &18.8 \\ \hline
LWS & 256/256 & 6.6M & 18.7 \\ \hline
\end{tabular}
\end{center}
\vspace{-0.1in}
\caption{Limited vs. Full Weight Sharing}
\vspace{-0.1in}
\label{table:lwsfws}
\end{table}

\subsection{Pooling Experiments}

Pooling is an important concept in CNNs which helps to reduce spectral variance in the input features. Similar to \cite{abdo:cnn}, we explore pooling in frequency only and not time, as this was shown to be optimal for speech. Because pooling can be dependent on the input sampling rate and speaking style, we compare the best pooling size for two different 50 hr tasks with different characteristics, namely 8kHZ speech - Switchboard Telephone Conversations (SWB) and 16kHz speech, English Broadcast News (BN).  Table \ref{table:pool} indicates that not only is pooling essential for CNNs, for all tasks pooling=3 is the optimal pooling size.  Note that we did not run the experiment with no pooling for BN, as it was already shown to not help for SWB.

\begin{table} [h!]
  \centering
  \begin{tabular}{|c|c|c|}
    \hline
    & WER-SWB &  WER- BN \\  \hline
    No pooling & 23.7 & -  \\ \hline
    pool=2 & 23.4 & 20.7\\ \hline
    pool=3 & 22.9 & 20.7\\ \hline
    pool=4 & 22.9 & 21.4\\ \hline
  \end{tabular}
  \caption{WER vs. pooling}\label{table:pool}
    \vspace{-0.15in}
\end{table}

\subsubsection{Type of Pooling}
Pooling is an important concept in CNNs which helps to reduce spectral variance in the input features. The work in \cite{tsainath:cnnLVCSR} explored using max pooling as the pooling strategy. Given a pooling region $R_j$ and a set of activations $\{a_1, \ldots a_{|R_j|}\} \in R_j$, the operation for max-pooling is shown in Equation \ref{eq:maxPool}.

\begin{equation}
s_j = \max_{i \in R_j} a_i
\label{eq:maxPool}
\end{equation}

One of the problems with max-pooling is that it can overfit the training data, and does not necessarily generalize to test data. Two pooling alternatives have been proposed to address some of the problems with max-pooling,  $l_p$ pooling \cite{sermanet:lp} and stochastic pooling \cite{fergus:sp}.

$l_p$ pooling looks to take a weighted average of activations $a_i$ in pooling region $R_j$, as shown in Equation \ref{eq:lp}. 
\begin{equation}
s_j = \left(\sum_{i \in R_j} a_i^p\right)^{\frac{1}{p}}
\label{eq:lp}
\end{equation}
$p=1$ can be seen as a simple form of averaging while $p=\infty$ corresponds to max-pooling. One of the problems with average pooling is that all elements in the pooling region are considered, so areas of low-activations may downweight areas of high activation. $l_p$ pooling for $p>1$ is seen as a tradeoff between average and max-pooling. $l_p$ pooling has shown to give large improvements in error rate in computer vision tasks compared to max pooling \cite{sermanet:lp}.

Stochastic pooling is another pooling strategy that addresses the issues of max and average pooling. In stochastic pooling, first a set of probabilities $p$ for each region $j$ is formed by normalizing the activations across that region, as shown in Equation \ref{eq:sp-prob}. 
\begin{equation}
p_i = \frac{a_i}{\sum_{k \in R_j} a_k}
\label{eq:sp-prob}
\end{equation}

\begin{equation}
s_j = a_l \texttt{ where } l \sim P(p_1, p_2, \ldots p_{|R_j|})
\label{eq:sp-pick}
\end{equation}

A multinomial distribution is created from the probabilities and the distribution is sampled based on $p$ to pick the location $l$ and corresponding pooled activation $a_l$. This is shown by Equation \ref{eq:sp-pick}. Stochastic pooling has the advantages of max-pooling but prevents overfitting due to the stochastic component. Stochastic pooling has also shown huge improvements in error rate in computer vision~\cite{fergus:sp}.

Given the success of $l_p$ and stochastic pooling, we compare both of these strategies to max-pooling on an LVCSR task. Results for the three pooling strategies are shown in Table \ref{table:poolingType}. Stochastic pooling seems to provide improvements over max and $l_p$ pooling, though the gains are slight. Unlike vision tasks, in appears that in tasks such as speech recognition which have a lot more data and thus better model estimates, generalization methods such as $l_p$ and stochastic pooling do not offer great improvements over max pooling.

\begin{table}[h!]
\begin{center}
\begin{tabular}{|c|c|} \hline
Method & WER  \\ \hline
Max Pooling & 18.9 \\ \hline
Stochastic Pooling & 18.8 \\ \hline
$l_p$ pooing & 18.9\\ \hline
\end{tabular}
\end{center}
\vspace{-0.1in}
\caption{Results with Different Pooling Types}
\vspace{-0.2in}
\label{table:poolingType}
\end{table}

\subsubsection{Overlapping Pooling}

The work presented in \cite{tsainath:cnnLVCSR} did not explore overlapping pooling in frequency. However, work in computer vision has shown that overlapping pooling can improve error rate by 0.3-0.5\% compared to non-overlapping pooling \cite{krizhevsky:imagenet}. One of the motivations of overlapping pooling is to prevent overfitting.

Table \ref{table:poolingStrat} compares overlapping and non-overlapping pooling on an LVCSR speech task. One thing to point out is that because overlapping pooling has many more activations, in order to keep the experiment fair, the number of parameters between non-overlapping and overlapping pooling was matched. The table shows that there is no difference in WER between overlapping or non-overlapping pooling. Again, on tasks with a lot of data such as speech, regularization mechanisms such as overlapping pooling, do not seem to help compared to smaller computer vision tasks.

\begin{table}[h!]
\begin{center}
\begin{tabular}{|c|c|} \hline
Method & WER  \\ \hline
Pooling No Overlap & 18.9 \\ \hline
Pooling with Overlap & 18.9 \\ \hline
\end{tabular}
\end{center}
\vspace{-0.1in}
\caption{Pooling With and Without Overlap}
\vspace{-0.1in}
\label{table:poolingStrat}
\end{table}

\subsubsection{Pooling in Time}
Most previous CNN work in speech explored pooling in frequency only (\cite{tsainath:cnnLVCSR}, \cite{abdo:cnn}, \cite{deng:hetPool}), though \cite{waibel1989phoneme} did investigate CNNs with pooling in time, but not frequency. However, most CNN work in vision performs pooling in both space and time \cite{sermanet:lp}, \cite{krizhevsky:imagenet}. In this paper, we do a deeper analysis of pooling in time for speech. One thing we must ensure with pooling in time in speech is that there is overlap between the pooling windows. Otherwise, pooling in time without overlap can be seen as subsampling the signal in time, which degrades performance. Pooling in time with overlap can thought of as a way to smooth out the signal in time, another form of regularization. 

Table \ref{table:poolTime} compares pooling in time for both max, stochastic and $l_p$ pooling. We see that pooling in time helps slightly with stochastic and $l_p$ pooling. However, the gains are not large, and are likely to be diminished after sequence training. It appears that for large tasks with more data, regularizations such as pooling in time are not helpful, similar to other regularization schemes such as $l_p$/stochastic pooling and pooling with overlap in frequency.

\begin{table}[h!]
\begin{center}
\begin{tabular}{|c|c|} \hline
Method & WER  \\ \hline
Baseline & 18.9 \\ \hline
Pooling in Time, Max & 18.9 \\ \hline
Pooling in Time, Stochastic &  18.8 \\ \hline
Pooling in Time, $l_p$ & 18.8 \\ \hline
\end{tabular}
\end{center}
\vspace{-0.1in}
\caption{Pooling in Time}
\vspace{-0.2in}
\label{table:poolTime}
\end{table}

\subsection{Incorporating Speaker-Adaptation into CNNs \label{sec:fmllr}}

In this section, we describe various techniques to incorporate speaker adapted features into CNNs. 

\subsubsection{fMLLR Features}

Since CNNs model correlation in time and frequency, they require the input feature space to have this property. This implies that commonly used feature spaces, such as Linear Discriminant Analysis, cannot be used with CNNs. In \cite{tsainath:cnnLVCSR}, it was shown that a good feature set for CNNs was VTLN-warped log-mel filter bank coefficients.

Feature-space maximum likelihood linear regression (fMLLR) \cite{gales:fmllr} is a popular speaker-adaptation technique used to reduce variability of speech due to different speakers. The fMLLR transformation applied to features assumes that either features are uncorrelated and can be modeled by diagonal covariance Gaussians, or features are correlated and can be modeled by a full covariance Gaussians.

While correlated features are better modeled by full-covariance Gaussians, full-covariance matrices dramatically increase the number of parameters per Gaussian component, oftentimes leading to parameter estimates which are not robust. 
Thus fMLLR is most commonly applied to a decorrelated space. When fMLLR was applied to the correlated log-mel feature space with a diagonal covariance assumption, little improvement in WER was observed \cite{tsainath:cnnLVCSR}.

Semi-tied covariance matrices (STCs) \cite{gales:stc} have been used to decorrelate the feature space so that it can be modeled by diagonal Gaussians. STC offers the added benefit in that it allows a few full covariance matrices to be shared over many distributions, while each distribution has its own diagonal covariance matrix.

In this paper, we explore applying fMLLR to correlated features (such as log-mel) by first decorrelating them such that we can appropriately use a diagonal Gaussian approximation with fMLLR. We then transform the fMLLR features back to the correlated space so that they can be used with CNNs.

The algorithm to do this is described as follows. First, starting from correlated feature space $\mathbf{f}$, we estimate an STC matrix $\mathbf{S}$ to map the features into an uncorrelated space. This mapping is given by transformation \ref{eq:stc}

\begin{equation}
\mathbf{S}\mathbf{f}
\label{eq:stc}
\end{equation}

Next, in the uncorrelated space, an fMLLR $\mathbf{M}$ matrix is estimated, and is applied to the STC transformed features. This is shown by transformation \ref{eq:fmllr}
\begin{equation}
\mathbf{M}\mathbf{S}\mathbf{f}
\label{eq:fmllr}
\end{equation}

Thus far, transformations  \ref{eq:stc} and \ref{eq:fmllr} demonstrate standard transformations in speech with STC and fMLLR matrices. However, in speech recognition tasks, once features are decorrelated with STC, further transformation (i.e. fMLLR, fBMMI) are applied in this decorrelated space, as shown in transformation \ref{eq:fmllr}. The features are never transformed back into the correlated space.

However for CNNs, using correlated features is critical. By multiplying the fMLLR transformed features by an inverse STC matrix, we can map the decorrelated fMLLR features back to the correlated space, so that they can be used with a CNN. The transformation we propose is given in transformation \ref{eq:invSTC}

\begin{equation}
\mathbf{S^{-1}}\mathbf{M}\mathbf{S}\mathbf{f}
\label{eq:invSTC}
\end{equation}

\subsection{Multi-scale CNN/DNNs}

The information captured in each layer of a neural network varies from more general to more specific concepts. For example, in speech  lower layers focus more on speaker adaptation and higher layers focus more on discrimination. In this section, we look to combine inputs from different layers of a neural network to explore if complementarity between different layers could potentially improve results further. This idea, known as multi-scale neural networks \cite{sermanet:lp} has been explored before for computer vision.

Specifically, we look at combining the output from 2 fully-connected and 2 convolutional layers. This output is fed into 4 more fully-connected layers, and the entire network is trained jointly. This can be thought of as combining features generated from a DNN-style and CNN-style network. Note for this experiment, the same input feature, (i.e., log-mel features) were used for both DNN and CNN streams. Results are shown in Table 5. A small gain is observed by combining DNN and CNN features, again much smaller than gains observed in computer vision. However, given that a small improvement comes at the cost of such a large parameter increase, and the same gains can be achieved by increasing feature maps in the CNN alone (see Table \ref{table:lwsfws}), we do not see huge value in this idea. It is possible however, that combining CNNs and DNNs with different types of input features which are complimentary, could potentially show more improvements.

\subsection{I-vectors}

\subsubsection{Results}

Results with the proposed fMLLR idea are shown in Table \ref{table:fmllr}. Notice that by applying fMLLR in a decorrelated space, we can achieve a 0.5\% improvement over the baseline VTLN-warped log-mel system. This gain was not possible in \cite{tsainath:cnnLVCSR} when fMLLR was applied directly to correlated log-mel features.

\begin{table}[h!]
\begin{center}
\begin{tabular}{|c|c|} \hline
Feature & WER  \\ \hline
VTLN-warped log-mel+d+dd &  18.8   \\  \hline
proposed fMLLR + VTLN-warped log-mel+d+dd & \textbf{18.3} \\ \hline
\end{tabular}
\end{center}
\vspace{-0.1 in}
\caption{WER With Improved fMLLR Features}
\label{table:fmllr}
\vspace{-0.2 in}
\end{table}

\subsection{Rectified Linear Units and Dropout \label{sec:relu}}

At IBM, two stages of Neural Network training are performed. First, DNNs are trained with a frame-discriminative stochastic gradient descent (SGD) cross-entropy (CE) criterion. Second, CE-trained DNN weights are re-adjusted using a sequence-level objective function \cite{bedk:nn}. Since speech is a sequence-level task, this objective is more appropriate for the speech recognition problem. Numerous studies have shown that sequence training provides an additional 10-15\% relative improvement over a CE trained DNN \cite{bedk:hf}, \cite{tsainath:cnnLVCSR}. Using a 2nd order Hessian-free (HF) optimization method is critical for performance gains with sequence training compared to SGD-style optimization, though not as important for CE-training \cite{bedk:hf}.

Rectified Linear Units (ReLU) and Dropout \cite{hinton:dropout} have recently been proposed as a way to regularize large neural networks. In fact, ReLU+dropout was shown to provide a 5\% relative reduction in WER for cross-entropy-trained DNNs on a 50-hr English Broadcast News LVCSR task \cite{gdahl:relu}. However, subsequent HF sequence training \cite{bedk:hf} \emph{that used no dropout} erased some of these gains, and performance was similar to a DNN trained with a sigmoid non-linearity and no dropout. Given the importance of sequence-training for neural networks, in this paper, we propose a strategy to make dropout effective during HF sequence training. Results are presented in the context of CNNs, though this algorithm can also be used with DNNs.

\subsubsection{Hessian-Free Training}

One popular 2nd order technique for DNNs is Hessian-free (HF) optimization~\cite{Martens2010}.  Let $\boldsymbol{\theta}$ denote the network parameters, $\mathcal{L}(\boldsymbol{\theta})$ denote a loss function, $\nabla \mathcal{L}(\boldsymbol{\theta})$ denote the gradient of the loss with respect to the parameters,
$\mathbf{d}$ denote a search direction, and $\mathbf{B(\boldsymbol{\theta})}$ denote a Hessian approximation matrix characterizing the curvature of the loss around $\boldsymbol{\theta}$.  The central idea in HF optimization is to iteratively form a quadratic approximation to the loss and to minimize this approximation using conjugate gradient (CG).
\begin{equation}
\mathcal{L}(\boldsymbol{\theta}+\mathbf{d}) \approx
\mathcal{L}(\boldsymbol{\theta}) + \nabla
\mathcal{L}(\boldsymbol{\theta})^{T} \mathbf{d} + \frac{1}{2}
\mathbf{d}^{T} \mathbf{B(\boldsymbol{\theta})} \mathbf{d}
\label{eq:hf}
\end{equation}

During each iteration of the HF algorithm, first, the gradient is computed using all training examples. Second, since the Hessian cannot be computed exactly, the curvature matrix  $\mathbf{B}$ is approximated by a damped version of the Gauss-Netwon matrix $\mathbf{G(\boldsymbol{\theta})} + \lambda \mathbf{I}$, where $\lambda$ is set via Levenberg-Marquardt. Then, Conjugate gradient (CG) is run for multiple-iterations until the relative per-iteration progress made in minimizing the CG objective function falls below a certain tolerance. During each CG iteration, Gauss-Newton matrix-vector products are computed over a sample of the training data. 


\subsubsection{Dropout}

Dropout is a popular technique to prevent over-fitting during neural network training \cite{hinton:dropout}. Specifically, during the feed-forward operation in neural network training, dropout omits each hidden unit randomly with probability $p$. This prevents complex co-adaptations between hidden units, forcing hidden units to not depend on other units. Specifically, using dropout the activation $\mathbf{y}^l$  at layer $l$ is given by Equation \ref{eq:dropout}, where $\mathbf{y}^{l-1}$ is the input into layer $l$, $\mathbf{W}^l$ is the weight for layer $l$, $\mathbf{b}$ is the bias, $f$ is the non-linear activation function (i.e. ReLU) and $\mathbf{r}$ is a binary mask, where each entry is drawn from a Bernoulli($p$) distribution with probability $p$ of being 1. Since dropout is not used during decoding, the factor $\frac{1}{1-p}$ used during training ensures that at test time, when no units are dropped out, the correct total input will reach each layer.

\begin{equation}
\mathbf{y}^l = f\left(\frac{1}{1-p} \mathbf{W}^{l}(\mathbf{r}^{l-1} * \mathbf{y}^{l-1})+\mathbf{b}^{l}\right)
\label{eq:dropout}
\end{equation}

\subsubsection{Combining HF + Dropout}

Conjugate gradient tries to minimize the quadratic objective function given in Equation \ref{eq:hf}. For each CG iteration, the damped Gauss-Netwon matrix, $\mathbf{G(\boldsymbol{\theta})}$, is estimated using a subset of the training data. This subset is fixed for all iterations of CG. This is because if the data used to estimate $\mathbf{G(\boldsymbol{\theta})}$ changes, we are no longer guaranteed to have conjugate search directions from iteration to iteration.

Recall that dropout produces a random binary mask for each presentation of each training instance. However, in order 
to guarantee good conjugate search directions, for a given utterance, the dropout mask per layer cannot change during CG. The appropriate way to incorporate dropout into HF is to allow the dropout mask to change for different layers and different utterances, but to fix it for all CG iterations while working with a specific layer and specific utterance (although the masks can be refreshed between HF iterations).

As the number of network parameters is large, saving out the dropout mask per utterance and layer is infeasible. Therefore, we randomly choose a seed for each utterance and layer and save this out. Using a randomize function with the same seed  guarantees that the same dropout mask is used per layer/per utterance.

\subsubsection{Results}

We experimentally confirm that using a dropout probability of $p=0.5$ in the 3rd and 4th layers is reasonable, and the dropout in all other layers is zero. For these experiments, we use 2K hidden units for the fully connected layers, as this was found to be more beneficial with dropout compared to 1K hidden units \cite{gdahl:relu}. 

\begin{figure}[h!]
\begin{center}
\includegraphics[width=2.5in]{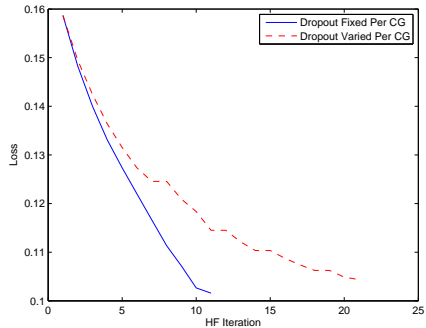}
\vspace{-0.1in}
\caption{Held-out Loss With Dropout Techniques}
\label{fig:hfLoss}
\end{center}
\vspace{-0.3in}
\end{figure}

Results with different dropout techniques are shown in Table \ref{table:reluHF}. Notice that if no dropout is used, the WER is the same as sigmoid, a result which was also found for DNNs in \cite{gdahl:relu}. By using dropout but fixing the dropout mask per utterance across all CG iterations, we can achieve a 0.6\% improvement in WER. Finally, if we compare this to varying the dropout mask per CG training iteration, the WER increases. Further investigation in Figure \ref{fig:hfLoss} shows that if we vary the dropout mask, there is slow convergence of the loss during training, particularly when the number of CG iterations increases during the later part of HF training. This shows experimental evidence that if the dropout mask is not fixed, we cannot guarantee that CG iterations produce conjugate search directions for the loss function. 

\begin{table}[h!]
\begin{center}
\begin{tabular}{|c|c|} \hline
Non-Linearity & WER  \\ \hline
Sigmoid & 15.7  \\ \hline
ReLU, No Dropout & 15.6 \\ \hline
ReLU, Dropout Fixed for CG Iterations & \textbf{15.0} \\ \hline
ReLU, Dropout Per CG Iteration & 15.3 \\ \hline
\end{tabular}
\end{center}
\vspace{-0.1 in}
\caption{WER of HF Sequence Training + Dropout}
\vspace{-0.1 in}
\label{table:reluHF}
\end{table}

Finally, we explore if we can reduce the number of CE iterations before moving to sequence training. A main advantage of sequence training is that it is more closely linked to the speech recognition objective function compared to cross-entropy. Using this fact, we explore how many iterations of CE are actually necessary before moving to HF training. Table \ref{table:reluHFIter} shows the WER for different CE iterations, and the corresponding WER after HF training. Note that HF training is started and lattices are dumped using the CE weight that is stopped at. Notice that just by annealing two times, we can achieve the same WER after HF training, compared to having the CE weights converge. This points to the fact that spending too much time in CE is unnecessary. Once the weights are in a relatively decent space, it is better to just jump to HF sequence training which is more closely matched to the speech objective function.

\begin{table}[h!]
\begin{center}
\begin{tabular}{|c|c|c|c|} \hline
CE Iter & \# Times Annealed & CE WER & HF WER  \\ \hline
4 & 1 & 20.8 & 15.3 \\ \hline
6 & 2 & 19.8 &15.0 \\ \hline
8 & 3 & 19.4 &15.0 \\ \hline
13& 7 & 18.8 &15.0 \\ \hline
\end{tabular}
\end{center}
\vspace{-0.1in}
\caption{HF Seq. Training WER Per CE Iteration}
\vspace{-0.3in}
\label{table:reluHFIter}
\end{table}

\section{Results}

In this section, we analyze CNN performance with the additions proposed in Section 3, namely fMLLR and ReLU + dropout. Results are shown on both a 50 and 400 hr English Broadcast News task. 
\subsection{50-hour English Broadcast News}

\subsubsection{Experimental Setup}

Following the setup in \cite{tsainath:cnnLVCSR}, the hybrid DNN is trained using speaker-adapted, VTLN+fMLLR features as input, with a context of 9 frames. A 5-layer DNN with 1,024 hidden units per layer and a sixth softmax layer with 2,220 output targets is used. All DNNs are pre-trained, followed by CE training and then HF sequence-training \cite{bedk:hf}. The DNN-based feature system is also trained with the same architecture, but uses 512 output targets. A PCA is applied on top of the DNN before softmax to reduce the dimensionality from 512 to 40. Using these DNN-based features, we apply maximum-likelihood GMM training, followed by feature and model-space discriminative training using the BMMI criterion. In order to fairly compare results to the DNN hybrid system, no MLLR is applied to the DNN feature-based system. The old CNN systems are trained with VTLN-warped log-mel+d+dd features, and a sigmoid non-linearity. The proposed CNN-based systems are trained with the fMLLR features described in Section \ref{sec:fmllr}, and ReLU+Dropout discussed in Section \ref{sec:relu}. 

\subsubsection{Results}
Table \ref{table:results50} shows the performance of proposed CNN-based feature and hybrid systems, and compares this to DNN and old CNN systems. The proposed CNN hybrid system offers between a 6-7\% relative improvement over the DNN hybrid, and a 2-3\% relative improvement over the old CNN hybrid system. While the proposed CNN-based feature system offers a modest 1\% improvement over the old CNN-based feature system, this slight improvements with feature-based system is not surprising all. We have observed huge relative improvements in WER (10-12\%) on a hybrid sequence trained DNN with 512 output targets, compared to a hybrid CE-trained DNN. However, after features are extracted from both systems, the gains diminish down to 1-2\% relative \cite{tsainath:dbnBN}. Feature-based systems use the neural network to learn a feature transformation, and seem to saturate in performance even when the hybrid system used to extract the features improves. Thus, as the table shows, there is more potential to improve a hybrid system as opposed to a feature-based system.

\begin{table} [h!]
  \centering
  \begin{tabular}{|c||c|c|}
    \hline
    model  & {\tt dev04f}& {\tt rt04} \\ \hline
    Hybrid DNN & 16.3 & 15.8 \\ \hline
    Old Hybrid CNN \cite{tsainath:cnnLVCSR} & 15.8 & 15.0\\ \hline
    Proposed Hybrid CNN & \textbf{15.4} & \textbf{14.7} \\ \hline \hline
   DNN-based Features & 17.4 & 16.6 \\ \hline 
    Old CNN-based Features \cite{tsainath:cnnLVCSR} & 15.5 & 15.2\\ \hline
    Proposed CNN-based Features &15.3 & 15.1 \\ \hline
  \end{tabular}
  \vspace{-0.05in}
  \caption{WER on Broadcast News, 50 hours}\label{table:results50}
  \vspace{-0.15in}
\end{table}

\subsection{400 hr English Broadcast News}

\subsubsection{Experimental Setup}

We explore scalability of the proposed techniques on 400 hours of English Broadcast News \cite{bedk:nn}. Development is done on the DARPA EARS {\tt  dev04f} set.  Testing is done on the DARPA EARS {\tt rt04} evaluation set. 
The DNN hybrid system uses fMLLR features, with a 9-frame context, and use five hidden layers each containing 1,024 sigmoidal units.  The DNN-based feature system is trained with 512 output targets, while the hybrid system has 5,999 output targets. Results are reported after HF sequence training. Again, the proposed CNN-based systems are trained with the fMLLR features described in Section \ref{sec:fmllr}, and ReLU+Dropout discussed in Section \ref{sec:relu}.

\subsubsection{Results}
Table \ref{table:results400} shows the performance of the proposed CNN system compared to DNNs and the old CNN system. While the proposed 512-hybrid CNN-based feature system did improve (14.1 WER) over the old CNN (14.8 WER), performance slightly deteriorates after CNN-based features are extracted from the network. However, the 5,999-hybrid CNN offers between a 13-16\% relative improvement over the DNN hybrid system, and between a  4-5\% relative improvement over the old CNN-based features systems. This helps to strengthen the hypothesis that hybrid CNNs have more potential for improvement, and the proposed fMLLR and ReLU+dropout techniques provide substantial improvements over DNNs and CNNs with a sigmoid non-linearity and VTLN-warped log-mel features. 
\begin{table} [h!]
  \centering
  \begin{tabular}{|c||c|c|}
    \hline
    model  & {\tt dev04f} & {\tt rt04}\\ \hline
    Hybrid DNN & 15.1 &  13.4\\ \hline
    DNN-based Features & 15.3 & 13.5\\ \hline
    Old CNN-based Features  \cite{tsainath:cnnLVCSR} & 13.4& 12.2\\ \hline
    Proposed CNN-based Features  & 13.6 & 12.5\\ \hline
    Proposed Hybrid CNN  & \textbf{12.7} & \textbf{11.7} \\ \hline
  \end{tabular}
  \caption{WER on Broadcast News, 400 hrs}\label{table:results400}
  \vspace{-0.25in}
\end{table}

\section{Conclusions \label{sec:conclusions}}

In this paper, we explored various strategies to improve CNN performance. We incorporated fMLLR into CNN features, and also made dropout effective after HF sequence training. We also explored various pooling and weight sharing techniques popular in computer vision, but found they did not offer improvements for LVCSR tasks. Overall, with the proposed fMLLR+dropout ideas, we were able to improve our previous best CNN results by 2-5\% relative.
\footnotesize
\bibliographystyle{IEEEbib}
\bibliography{IEEEabrv,main}

\begin{thebibliography}{10}

\bibitem{Hinton12}
G.~Hinton, L.~Deng, D.~Yu, G.~Dahl, A.~Mohamed, N.~Jaitly, A.~Senior,
  V.~Vanhoucke, P.~Nguyen, T.~N. Sainath, and B.~Kingsbury,
\newblock ``{Deep Neural Networks for Acoustic Modeling in Speech
  Recognition},''
\newblock {\em IEEE Signal Processing Magazine}, vol. 29, no. 6, pp. 82--97,
  2012.

\bibitem{lecun:cnn}
Y.~LeCun and Y.~Bengio,
\newblock ``{Convolutional Networks for Images, Speech, and Time-series},''
\newblock in {\em The Handbook of Brain Theory and Neural Networks}. MIT Press,
  1995.

\bibitem{abdo:cnn}
O.~Abdel-Hamid, A.~Mohamed, H.~Jiang, and G.~Penn,
\newblock ``{Applying Convolutional Neural Network Concepts to Hybrid NN-HMM
  Model for Speech Recognition},''
\newblock in {\em Proc. ICASSP}, 2012.

\bibitem{tsainath:cnnLVCSR}
T.N. Sainath, A.~Mohamed, B.~Kingsbury, and B.~Ramabhadran,
\newblock ``{Deep Convolutional Neural Networks for LVCSR},''
\newblock in {\em {Proc. ICASSP}}, 2013.

\bibitem{deng:hetPool}
L.~Deng, O.~Abdel-Hamid, and D.~Yu,
\newblock ``{A Deep Convolutional Neural Network using Heterogeneous Pooling
  for Trading Acoustic Invariance with Phonetic Confusion},''
\newblock in {\em Proc. ICASSP}, 2013.

\bibitem{sermanet:lp}
P.~Sermanet, S.~Chintala, and Y.~LeCun,
\newblock ``{Convolutional neural networks applied to house numbers digit
  classification},''
\newblock in {\em Pattern Recognition (ICPR), 2012 21st International
  Conference on}, 2012.

\bibitem{fergus:sp}
M.~Zeiler and R.~Fergus,
\newblock ``{Stochastic Pooling for Regularization of Deep Convolutional Neural
  Networks},''
\newblock in {\em Proc. of the International Conference on Representaiton
  Learning (ICLR)}, 2013.

\bibitem{krizhevsky:imagenet}
A.~Krizhevsky, I.~Sutskever, and G.~Hinton,
\newblock ``{Imagenet Classification with Deep Convolutional Neural
  Networks},''
\newblock in {\em Advances in Neural Information Processing Systems}, 2012.

\bibitem{lecun:cnnVision}
Y.~LeCun, F.~Huang, and L.~Bottou,
\newblock ``{Learning Methods for Generic Object Recognition with Invariance to
  Pose and Lighting},''
\newblock in {\em Proc. CVPR}, 2004.

\bibitem{gales:fmllr}
M.J.F. Gales,
\newblock ``{Maximum likelihood linear transformations for HMM-based Speech
  Recognition},''
\newblock {\em Computer Speech and Language}, vol. 12, no. 2, pp. 75--98, 1998.

\bibitem{bedk:hf}
B.~Kingsbury, T.~N. Sainath, and H.~Soltau,
\newblock ``{Scalable Minimum Bayes Risk Training of Deep Neural Network
  Acoustic Models Using Distributed Hessian-free Optimization},''
\newblock in {\em Proc. Interspeech}, 2012.

\bibitem{gdahl:relu}
G.E. Dahl, T.N. Sainath, and G.E. Hinton,
\newblock ``{Improving Deep Neural Networks for LVCSR Using Rectified Linear
  Units and Dropout},''
\newblock in {\em {Proc. ICASSP}}, 2013.

\bibitem{waibel1989phoneme}
A.~Waibel, T.~Hanazawa, G.~Hinton, K.~Shikano, and K.J Lang,
\newblock ``{Phoneme Recognition using Time-delay Neural Networks},''
\newblock {\em IEEE Transactions on Acoustics, Speech and Signal Processing},
  vol. 37, no. 3, pp. 328--339, 1989.

\bibitem{gales:stc}
M.J.F. Gales,
\newblock ``{Semi-tied Covariance Matrices for Hidden Markov Models},''
\newblock {\em IEEE Transactions on Speech and Audio Processing}, vol. 7, pp.
  272--281, 1999.

\bibitem{bedk:nn}
B.~Kingsbury,
\newblock ``{Lattice-Based Optimization of Sequence Classification Criteria for
  Neural-Network Acoustic Modeling},''
\newblock in {\em Proc. ICASSP}, 2009.

\bibitem{hinton:dropout}
G.E. Hinton, N.~Srivastava, A.~Krizhevsky, I.~Sutskever, and R.~Salakhutdinov,
\newblock ``{Improving Neural Networks by Preventing Co-Adaptation of Feature
  Detectors},''
\newblock {\em The Computing Research Repository (CoRR)}, vol. 1207.0580, 2012.

\bibitem{Martens2010}
J.~Martens,
\newblock ``Deep learning via {Hessian-free} optimization,''
\newblock in {\em Proc. Intl. Conf. on Machine Learning (ICML)}, 2010.

\bibitem{tsainath:dbnBN}
T.~N. Sainath, B.~Kingsbury, and B.~Ramabhadran,
\newblock ``{Auto-Encoder Bottleneck Features Using Deep Belief Networks},''
\newblock in {\em Proc. ICASSP}, 2012.

\end{thebibliography}
\end{document}